# VizInspect Pro - Automated Optical Inspection (AOI) solution

Faraz Waseem, Mahesh Sankaran, Sanjit Menon, Srinath Cheluvaraja, Haotian Xu, Debashis Mondal,
Anand Kumar, Arun Kumar, Fengmei Liu, Li Yang, Yang song, Ashutosh Malgaonkar, Rakesh Rai
Hennadka, Omkar Venkata Siva, Chetan Kumar, Evgueni Gordienko

## 1 ABSTRACT

Traditional vision based Automated Optical Inspection (referred to as AOI in paper) systems present multiple challenges in factory settings including inability to scale across multiple product lines, requirement of vendor programming expertise, little tolerance to variations and lack of cloud connectivity for aggregated insights. The lack of flexibility in these systems presents a unique opportunity for a deep learning based AOI system specifically for factory automation.

The proposed solution, VizInspect pro is a generic computer vision based AOI solution built on top of Leo - An edge AI platform. Innovative features that overcome challenges of traditional vision systems include deep learning based image analysis which combines the power of self-learning with high speed and accuracy, an intuitive user interface to configure inspection profiles in minutes without ML or vision expertise and the ability to solve complex inspection challenges while being tolerant to deviations and unpredictable defects.

This solution has been validated by multiple external enterprise customers with confirmed value propositions. In this paper we show you how this solution and platform solved problems around model development, deployment, scaling multiple inferences and visualizations.

## 2 INTRODUCTION

VizInspectPro is a deep learning based application designed specifically for manufacturing automation to overcome the challenges of traditional vision based AOI systems. VizInspectPro's innovative features around simplified training, intuitive user interface, ease of setup and unmatched scalability allow enterprises with extensive manufacturing assembly lines to easily identify defects across their product range using cameras and computer vision (CV) models in real time. This solution generates actionable insights for Quality Inspectors to understand defect distribution and initiate corrective actions. speed and flexibility to create new or existing products in greater volumes.

VizInspectPro uses Leo - the edge AI platform[12] to provide real time actionable insights. This AI platform has the ability to provide ML training, inferencing and feedback workflows to solutions like VizInspectPro. This allows non data scientists, in this case plant operators to train against seed models using an in line model training pipeline. The same operators can then enable real time inferencing pipelines that allow for real time inspections and the ability to provide feedback to models to allow for further refinement by retraining.

The inherent data ingestion, data processing, ML training and inferencing, data storage and data visualization aspects of Leo allow for multiple solutions to be built on top of the platform. While the inner workings of the Leo platform are not in the scope of this paper, a paper on the Leo platform was presented in 2020 tech pulse.

In this paper we'll demonstrate the features of VizInspectPro deployed on edge compute infrastructure, the ability to create multiple IOT and CV solutions using building blocks available on the Leo Platform and go into detail around the ML models and techniques we have used to ensure accuracy and efficiency.The results show that by employing the proposed method, inspection volumes can be reduced significantly and thus economic advantages can be generated.

Analyzing large amounts of data based on complex machine learning algorithms requires significant computational capabilities. Therefore, much processing of data takes place in on-premises data centers or cloud-based infrastructure. However, the arrival of edge computing has given rise to the era of deploying advanced machine learning methods such as convolutional neural networks, or CNNs, at the edges of the network for "edge-based" ML. Deep learning applications like VizInspectPro running on the edge need a platform as a service (PaaS) layer that will allow for ingestion of data in real time in order to provide actionable data and ML insights.. Our Leo - edge AI platform working as a PaaS layer provides multiple capabilities to ingest, process, store and visualize data for SaaS applications like VizInspectPro.

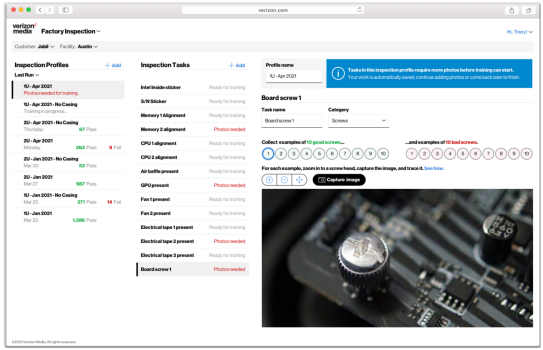
*VizInspectPro Model Training UI*

## 3 BACKGROUND

### 3.1 Vision

Defect free, high quality products are a critical success factor for the long-term competitiveness of manufacturing companies. Despite the increasing challenges of rising product variety and complexity and the necessity of economic manufacturing, a comprehensive and reliable quality inspection is often indispensable. In consequence, high inspection volumes turn inspection processes into manufacturing bottlenecks.

In this contribution, we propose a new integrated solution of automated optical inspection (AOI) for industrial automation by utilizing Machine Learning techniques and Edge Cloud Computing technology. In contrast to state-of-the-art contributions, we propose a holistic approach comprising the target-oriented data acquisition and processing, modelling and model deployment as well as the technological implementation in an existing IT plant infrastructure. A real industrial use case in smart manufacturing is presented to underline the procedure and benefits of the proposed method.

### 3.2 Value Proposition

Traditional vision based AOI systems used in manufacturing industries for defect detection have multiple challenges that primarily inhibit scaling across multiple product lines. Current solutions in the market today cannot be scaled across different products without heavy dollar investments due to their lack of flexibility. Programming a robot and vision systems requires deep expertise and niche skills that are not readily available within an enterprise. This involves the use of vendor resources which costs both time and money. Additionally industrial robot reconfiguration is difficult. Altering permanent installation structures is difficult and unsafe. Other challenges include the lack of cloud integration for transfer of images and historical analytics.

VizInspectPro overcomes these challenges by providing several advantages over traditional vision based systems. These include a) Ability to create inspection profiles by simply uploading good and bad samples b) Annotate images though user intuitive UI c) One click training initiation without any Data Science or vision expertise d) Create as many profiles needed to cover entire product line (Scalability) e) One click Inspection Initiation f) Defect review capability with highlighted defect region g) Higher tolerance to natural variations and lighting h) Capability of performing complex inspection tasks

### 3.2 Market and Competitive Analysis

The automated optical inspection market was valued at USD 598 million in 2020 and is projected to reach USD **1.7B** by 2026. It is expected to grow at a CAGR of 20.8% during the forecast period. Advantages of AOI over other inspection methods, upsurge in the demand for consumer electronics amidst pandemic, rising need for miniature, high-speed PCBs, demand for higher productivity by electronics manufacturing services (EMS) companies, and growing demand for electronics in automotive sector are contributing to the growth of the automated optical inspection market. Advent of SMART technology, newer applications of AOI systems apart from PCB inspection, and growing demand for AOI systems for inspection of IC substrates act as a growth opportunity for the market players.

Koh Young (South Korea), Test Research, Inc. (TRI) (Taiwan), Omron (Japan), Camtek (Israel), Viscom (Germany), Saki Corporation (Japan), Nordson (US), KLA (US), Cyberoptics (US), and Goepel Electronics (Germany) are among the major players in the traditional vision based optical inspection market.

In the MLbased AOI market there are very few players. IBM is a nascent player in the market with their Maximo visual inspection solution but that product is still in its infancy. Traditional vision based AOI manufacturers are also trying to build new age alternatives leveraging deep learning & computer vision.

## 4 SOLUTION ARCHITECTURE

The Viz InspectPro Solution is based on Project Leo - a Kubernetes based edge AI/ML analytics platform with components that also allow rich ingestion and integration. This simplified the usage and deployment stack upfront.

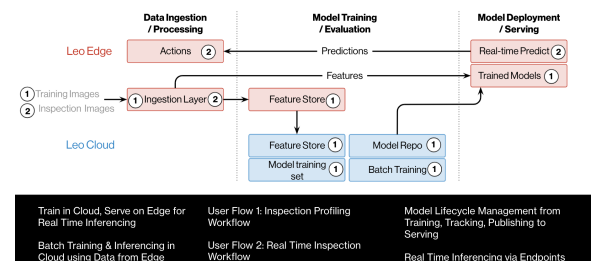
*VizInspectPro Solution Architecture*

Deployed on top of Leo is the Viz inspect pro designed to cater to AOI Solutions which are Camera facing automated real time inspection scenarios. The end user is a factory floor supervisor or employee who defines the product inspection profile with various inspection scenarios for detecting defects in assembled products. The complexity also lies in multiple cameras nearly simultaneously capturing the images on the assembly line.

The workflow is divided in to the following phases :
1. Profile and task/service  definition
2. Image acquisition
3. Inference workflow
4. Model Training (online and offline)
5. Retraining Feedback loop

## 4.1 Profile and task/service  definition

An inspection *profile* is a set of inspection tasks for a particular product which need to be evaluated in real-time during the inspection process. Each inspection area of the product to be examined is called an inspection task. An example of a real world task can be identifying presence/absence, examining proper  alignment, detecting irregularity or flaw etc..

Through a visually rich and advanced UI (web), original images aka. Golden images can be uploaded and regions where inspections need to be conducted are marked. Here we associate images for each task with good and bad examples.

We now have captured a profile, its tasks and their respective associated images in the system.

## 4.2 Image acquisition

Image acquisition is a critical part of any Optical Inspection system. At the time of writing of this paper, we are still evaluating various fixed multi-camera options. There might be a future need for spatial rotation of the camera using a robotic arm along different axes for capturing images from various angles. We designed our product to make it possible to extend the framework to fit robotic arm rotations too. The solution will use industrial grade cameras built specifically for performing optical inspection. We plan to add a plugin that dispatches newly clicked images from multiple cameras and POST them to a defined REST endpoint to begin our processing.

## 4.3  ML Training Platform (Offline and online)

Our model training platform can be conceptually divided into offline offline/dev and training/production environments. ML Training platform is based on the concept of End to End ML Training. It is based on KubeFlow which provides all the bells and whistles of modern machine learning training framework like running single server and distributed training workflows, experiment tracking, data ingestion, hyperparameter tuning and production level serving on Kubernetes ecosystem.[1]. It has a rich ecosystem of  third party frameworks supporting a wide variety of features like feature store,  ( via Apache feast), model management (modelDB) etc.[2]

### 4.3.1 Offline Model Training/Experimentation

In addition to the KubeFlow based ecosystem, Leo ML platform provides additional components to sync data between training environments like notebook/pipelines and persistence services like minio. We also have capability to deploy models trained outside our platform(like sagemaker)    to model registry in Leo ML Cloud or  in Leo Edge Deployment controller. We have a UI wizard which is user friendly to create new notebooks and generate data sync between minio and notebooks.

### 4.3.2  Online Model Training/Retraining.

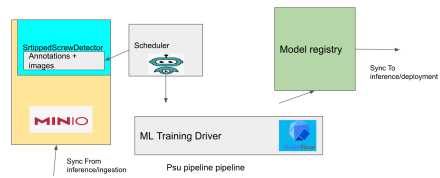

The main components of online model training
- Training scheduler,
- ML Training Driver
- Model registry.

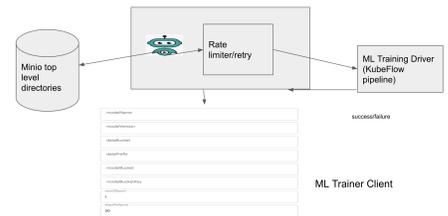

#### 4.3.2.1 Training Scheduler

Training Scheduler is an event listener of storage architecture (implemented with minio). When new data from the inference/ingestion platform lands in minio buckets along with marker file, it calls ML Training Driver with hyperparameters needed to run training pipeline. Training Schedulers also take care of retry logic,rate limiting of training requests and versioning of models.  The same design works for first time training of model or retraining of model with updated data.

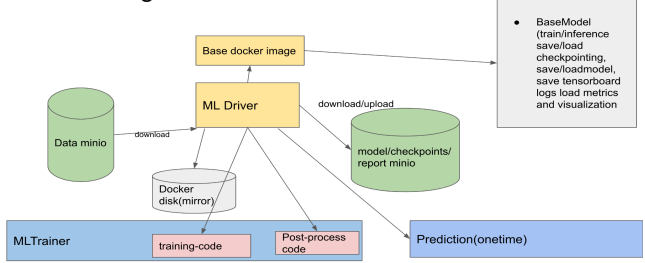

ML Training Driver

**4.3.2.2 Leo ML Training Driver**
Leo ML Training Driver is an orchestrator which can run a training pipeline for any model framework like tensorflow/pytorch/mxnet using configuration and base docker image ( packaged dependencies) and handle training code. It has utility methods of downloading/uploading data to minio. It also has utility methods to load/save checkpoints, tensor logs and model metrics and methods to save models. Leo ML Trainer is integrated with Kube Flow experiments and provides a dashboard to view training results. Leo ML Trainer can train a fresh model from new data samples or can start from a seed model from existing model checkpoints and updated data.

*VizInspectPro Model Training UI*

The ML Training driver is based on Kube Flow pipelines. We can orchestrate execution of multiple docker containers as DAG. The design pattern we are following consists of a "pre-process step" which downloads data from minio and does data transformation and cleanup. The training step runs a training loop for the model and periodically uploads checkpoints/model logs to minio. The post process step runs model on validation set and create reports and visualization and save it minio bucket for later visualization and inspection

**4.3.2.3 Model Registry**

ML Training Driver saves the model in the model registry which is implemented using minio. Model registry maintains staging and production environments. We have a UI based interface where we can see training statistics of different models.

*VizInspectPro Model Deployment UI*

Leo ML platform supports CICD workflow like manual or accuracy based promotions of models from staging to production environments. Production environment supports integration via adapters for model deployment on Edge and any cloud based platform. Model registry stores reference to docker containers for pre-processing, predict and post processing logic. These docker containers can be deployed as model-as-service to any platform which supports installation of docker containers like Seldon core.

The seamless rolling model version updates on the edge platform are a highlight of the platform along with model metrics, monitoring and auditing capabilities. Advanced model deployment capabilities like inference graphs for complex model serving, outliers, explainers, model drift, A/B testing, canary testing and multi-arm bandits are available within the edge platform.

## 4.4 Inference workflow

Once the camera system does its job of handing off images, there starts the workflow for inferencing workflows. Here is a summary of the workflow:

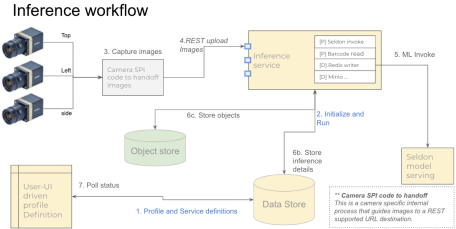

There is a lot of activity starting from barcode scanner, product image detection, alignment and cropping, ML Model inferencing and finally managing the entire workflow and its outputs. All this is done in this layer. MinIo is the Object storage as used in Leo and Redis is the cache used for storing the source of truth around profile data, paths etc.

From the image views sets of tasks are identified based on the regions matches for each task based on the profile. This generates cropped segment images from the main image for each camera view. After all the Task set identification and their corresponding ML model scoring, all outputs are captured and stored on to the Object store in a sensible layout. Later all paths and analytics are stored in Redis for any analytics lookups.

UI will offer drill down, filter and audit of the whole workflow. When an anomaly observation is missed then the user can re-trigger the model (re)training by adding the missing bad images to the set. This retraining process is explained in 4.5

## 4.5 Retraining Feedback loop

Retraining Feedback loop is a mechanism to accommodate images that are missing in the training process. Users get the opportunity to retrain the model that has missing labeled images.

A Rich and advanced UI visualization guides users to label the images(that are already captured in the previous camera stream) and add them to the missing set of images both bad and good. Then retraining will have to re-deploy after the model generation is completed. The solution completes the end-to-end flow from submitting missing images to retraining models and finally redeployment back to the production line. The beauty is that all this promotion process of new model deployment is happening transparently in the production line automatically. The retraining process hooks back to the same Kubeflow mechanics as described already in the previous sections.

## 5 USE CASES

### 5.1 Potential Edge Use Cases

This edge platform can be used to host a large number of edge use cases across multiple industries like manufacturing, oil and gas, mining, transportation, power and water, renewable energy, health care, retail, smart buildings, smart cities, and connected vehicles.

There are many classes of use cases. One is a real time streaming use case involving streaming data and examples of these include anomaly detection, predictive maintenance, cognitive video etc. The other is an on-demand use case involving batches of data and examples of these include smart logistics, location services , personalization systems etc.

### 5.2 Leo VizInspect Pro

Computer vision has been used in motherboard inspection to detect defects like loose cable, bad fan connection ,scratch screws and open dimms. One drawback of existing computer vision based defect detection systems is that they need custom model/processing for each kind of part and it takes time and resources to train new models for new server types. Also existing computer vision systems need professional grade expensive hardware. One of our clients who is a major computer server manufacturer is using cameras mounted on robotic arms to perform traditional visual based inspection.

This system needs to be recalibrated and reprogrammed for each new server type. Deep learning is promising as we can use models like segmentation and keypoint detection as building block and then create a generic system which can be trained to detect new defects by uploading/annotation good and bad examples by technicians without needing a data scientist on the factory floor and can work with commodity hardware like iphones.

These are few generic use cases we are solving for our client
- standoffs are positioned properly in keyholes.
- cables/connectors are plugged in the correct connector.
- cable routing is correct
- cables are properly plugged.
- Fan airflows are the right direction.
- dimm latches are all closed.
- M2 card properly connected and seated

**5.2.1 Data Collection Process:**
Client has not shared actual images of good and bad servers due to the confidential nature of these servers. They have pointed to similar servers and we collected images in the Sunnyvale server lab and manually created the defects. Also one of the hard requirements for this solution was to work with a small amount of training data like maximum 50 good and bad images for each part due to the laborious nature of data labeling and annotation at factory floor. We experimented with seed models by using images collected from server labs and the actual models will be retained on fly at the factory floor at client site. We collected images from a few server models and collected top, right and left views. Here are a few examples of data collection.

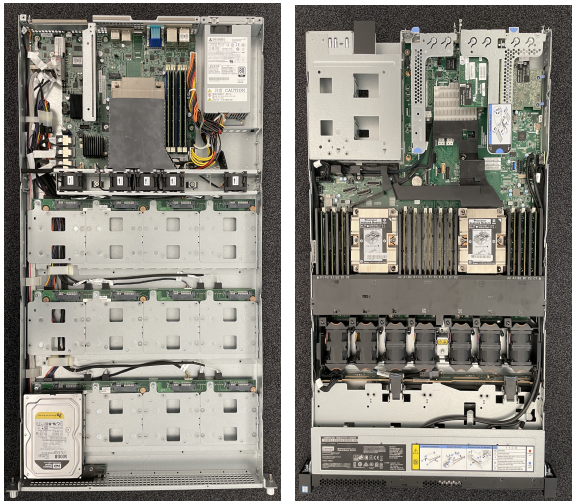

*1U Server Board - Top View*

**5.2.2 Image Alignment and Cropping**

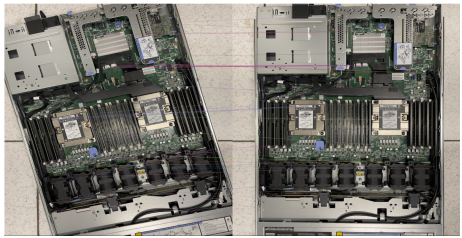

We used a divide and conquer approach to find defects. Instead of one model for a full server, we cropped the server into smaller parts each containing a specific target area. We used image alignment based on ORB and RANSAC algorithms. ORB stands for ORiented Fast and Rotated Brief [5] and RANSAC stands for Random Sample Consensus. Here are the results of the cropping of the power cable. Image at left is the test image after alignment and cropping, the image at center is the golden image after cropping and the image at right is if we don't use alignment. As per client specs we have to account for both displacement and rotation in target servers on the belt.

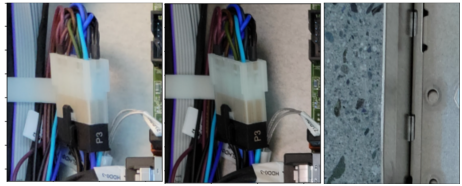

The cropped images are used for training each specific use case. Same process of alignment and cropping is repeated at inference time.

5.2.3 **Training Process**
Deep learning based Keypoint detection and segmentation are basic building blocks which are used to detect different defects. We will dig deep into the problem of detecting if the power cable is seated properly for demonstration of approach. Here are 4 different examples of good power cable and power cable partially unseated, power cable fully unseated and power cable borderline unseated in order from left to right and top to bottom.

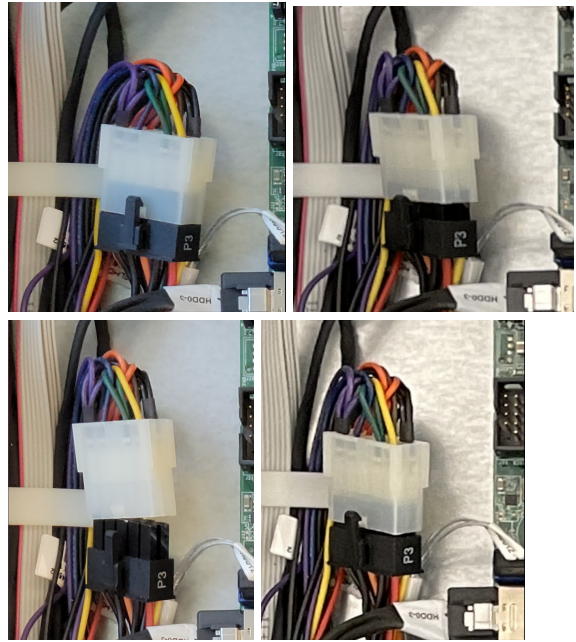

We used a label studio[16] for creating segmentation masks for training the model. The model was trained on 82 samples collected from the lab. We don't want to use hundreds or thousands of images for training as the actual model needs to be retrained on the factory floor on new data with a hard limit of 100 images from the client. We used image augmentation techniques(rotation, shift etc) to augment test data 10X times. The idea was to create a segmentation mask for the upper and lower part of the connector and identify the area of intersection between two connectors. Based on the intersection area, we derived a threshold based decision boundary to decide between positive and negative examples. The dark red area in diagrams below is the intersection between upper mask and lower mask.

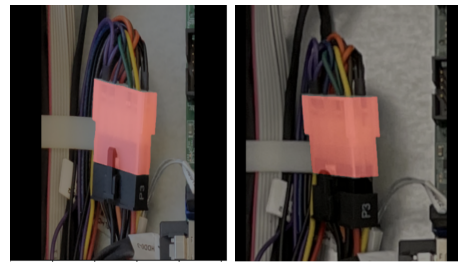

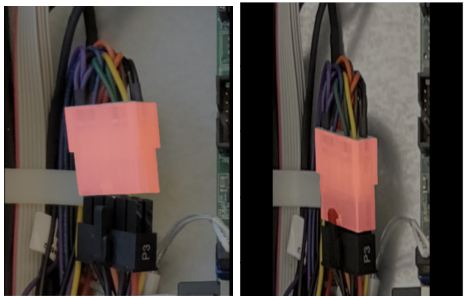

**UNet Model for segmentation**

For segmentation we used the Efficient-Unet as backbone model[3] with 31 millions parameters for our power cable case, Input resolution of images was 512*512*3. The model has been trained for 20 epochs with 200 steps per iteration and batch size of 8 images. We used 2 V100 Nvidia GPUs for the training loop.

**5.2.4 Experiment Results for Power Cable seating**
For power cable segmentation we have 160 images as validation sets. This is how distribution looks like for the kind of unseating problems we have.

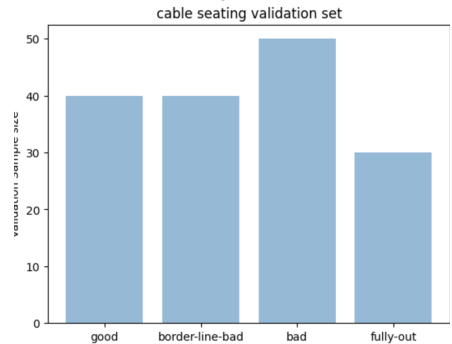

This is how the average intersection area looks like from the validation dataset.

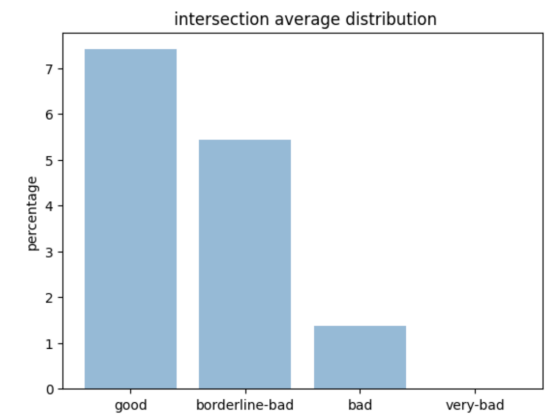

Using the intersection threshold of 5.7 derived from the training data, the model was able to correctly identify all different kinds of cable unseating problem correctly. preliminary results show **F1** of **100**% and **0**% False negative. In the actual production environment the accuracy might not be 100% as compared to testing in our lab environment. The minimum intersection we got for a good cable case was 5.95 which still has a margin with a decision boundary of 5.7.

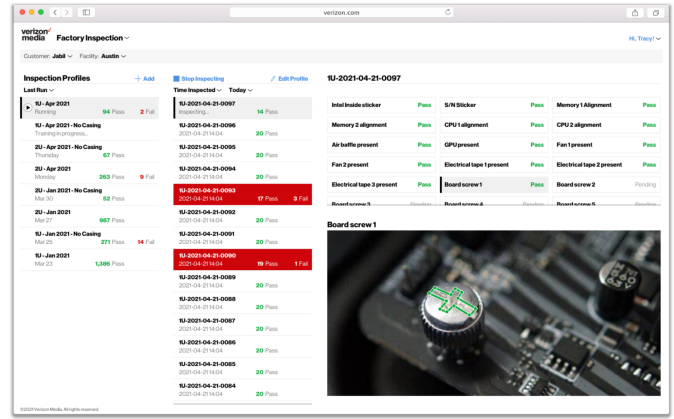

*VizInspectPro Inspection UI*

### 5.2.5 Cable Connector Detection

A template matching-based method is adopted to detect the presence of cable connectors. Since the cable connector detector only needs to recognize two status (cable is present or missing), we only use two kinds of templates, i.e., normal cable connector templates and background templates. In production, users are able to dynamically update the template library to gradually improve the model performance.

In order to reduce the search region of the sliding window, we predefine the candidate regions in the golden template, and crop the aligned test images according to these regions. The illumination changes may cause worse performance of the template matching method, so we make color alignment between the test images and the golden images following [].

We first calculate the similarity score between each test image and the templates, and then select the candidate region with the highest similarity score as the optimal connector region according to the K nearest neighbors (KNN) algorithm. The classification probability of the test image is estimated according to its similarity scores to the top K matched templates, with the contribution/weights from each template equal to the corresponding similarity score. This means that the closer nearest neighbor has greater effect for the classification category. Finally, the predicted probability is compared to a predefined threshold value, to generate the final binary prediction. The threshold is set at 50% probability by default, but the users are allowed to fine-tune this threshold to better fit their business needs (e.g. increase threshold to avoid type I error, or decrease the threshold to reduce type II error).

Our proposed template matching-based classification method is a metric learning-based supervised method, and it does not need a training process, which only requires a very small number of annotated images. Compared with the more advanced ML/DL method, our proposed method is a very practical solution for this task. Preliminary results show 100% accuracy in cross-validation tests with ~80 good templates and ~50 defective templates.

**5.2.6 Fan and Dimm Latches Experiment Result**

The fan airflow direction detection use case is to identify whether any of the airflow arrows in the fan stickers are pointing to the wrong direction. Two methodologies have been experimented for the fan airflow direction identification use case. Traditional computer vision method using OpenCV arrow shape detection and deep learning method using segmentation and key-point detection. We have chosen the vanilla key-point detection model after tests and iterations considering robustness and accuracy.
Similarly, multiple segmentation models have been experimented for the dimm latch open/close identification use case. The final selected model is the multi-class segmentation model using Efficient-Unet backbones.

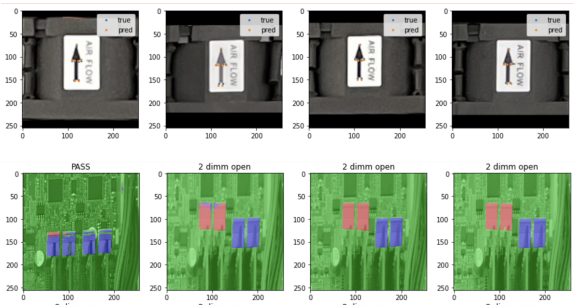

Fig. Result of use case experiment - top: fan airflow detection using key-point detection model; bottom - dimm latch open/close detection using multi-class segmentation model

**5.2.7 M2 card seated correctly or incorrectly**

The M2 card, also known as solid state hard drive, is increasingly being deployed in the latest servers because of its compact form factor and superior performance. Correct seating of the m2 card can be checked by similar methodologies as discussed above. Segmentation and keypoint detection methods have given 100% accuracy on some test examples using only very small training sets. Attached below is an example where m2 card connection is flagged as improper when its connector is exposed instead of being in its designated spot. The segmentation approach is able to detect this exposed connection (see red region on figure) even when a distracting object ( in this case a human hand) is present.

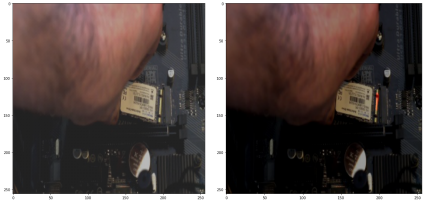

Fig. M2 card on left, and the result of m2 card being flagged on right by segmentation method.The red marks on connector are the prediction of segmentation method.

## 5 FUTURE

At the time of writing this paper, inspection use cases are based on the requirements from a specific Enterprise Computing OEM customer. The product vision is about building a generic inspection solution that can cater to manufacturing automation for different assembly/product lines. The solution can not only be used for inspecting the final product but can also be part of the supply chain where it can inspect various vendor supplied components. More inspection categories and related inspection areas will be added in future releases to broaden customer reach and overall inspection capabilities. Initial customer feedback will be used to improve user experience and model performance. We plan to add support for deployments on both Edge and Cloud and their interplay. We are also planning to add active learning for future development so we can guide our clients in the retraining process. Another dimension we are exploring is unsupervised learning strategies like autoencoders/GAN and active learning. We don't have enough unlabeled data to validate unsupervised learning but once we deploy our system at an actual factory we will get access to a larger amount of unlabeled data and we can validate unsupervised approaches. That will alleviate the need for manual annotation and labeling.

# 6 CONCLUSION

In this paper we have examined a novel AOI solution using deep learning techniques to overcome some of the challenges of traditional vision systems. This solution running on top of Leo edge AI platform allows for simple model training that can be done by non data scientists, real time model inferences and innovative ways to provide model feedback and retrain models.VizInspect Pro deployed on edge computing nodes can help enterprises increase inspection volumes thus overall manufacturing yield with greater speed and accuracy compared to human inspection.